\documentclass{article}
\usepackage[dblblindworkshop,final]{neurips_2026}
\workshoptitle{AXIOM Foundations of Efficient Deep Learning}
\usepackage[utf8]{inputenc}
\usepackage[T1]{fontenc}
\usepackage{array}
\usepackage{float}
\usepackage{hyperref}
\usepackage{url}
\usepackage{booktabs}
\usepackage{amsfonts}
\usepackage{nicefrac}
\usepackage{microtype}
\usepackage{graphicx}

\title{ICICLE: In-Context Inference for Curve-Learning Extrapolation}
\author{%
  Laurits Fredsgaard \quad Michael Riis Andersen \quad Mikkel N. Schmidt \\
  Department of Applied Mathematics and Computer Science \\
  Technical University of Denmark \\
  \texttt{laula@dtu.dk, miri@dtu.dk, mnsc@dtu.dk}
}

\begin{document}

\maketitle

\begin{abstract}
Freeze--thaw Bayesian optimization allocates training budget by predicting how
partially observed learning curves will continue. LC-PFN first applied prior-data fitted
networks (PFNs) to learning-curve extrapolation, and FT-PFN extended this approach to
configuration-aware freeze--thaw optimization with a ten-slot hyperparameter encoder.
We introduce In-Context Inference for Curve-Learning Extrapolation (ICICLE), an in-context
surrogate that accepts a variable number of hyperparameters through a shared set encoder and
is trained on a synthetic prior covering typed hyperparameters, heavy-tailed noise,
quantization, and divergence. ICICLE has 1.28 million parameters (8.7\% of FT-PFN's count) and produces similar mean-regret
trajectories to FT-PFN on LCBench, PD1, and TaskSet. On the 86-task Meta-Album benchmark with
33 hyperparameters, ICICLE reaches lower mean normalized regret than Quick-Tune overall and
on extended, with no clear difference on mini or micro. It does so without pre-training or meta-training on Meta-Album tasks,
whereas Quick-Tune is meta-trained on related Meta-Album tasks. Our release samples synthetic
training tasks on the fly and provides a freeze--thaw scheduler for Ray Tune and an ONNX model
for inference without JAX or PyTorch.
\end{abstract}

\section{Introduction}\label{sec:intro}

Random search improves on grid search~\citep{bergstra2012random} and parallelizes easily:
with enough workers, its wall-clock cost approaches that of a single training run. However,
it spends substantial compute on poor configurations because every run is trained to
completion. The asynchronous successive halving algorithm
(ASHA)~\citep{jamieson2016nonstochastic,li2020asha} and
Hyperband~\citep{li2018hyperband} stop poor runs early and keep the parallelism.
Their decisions depend on performance observed so far, so they can stop late bloomers or
continue runs that later plateau or become unstable. Conventional Bayesian optimization
selects configurations adaptively, but
observes their performance only after full training, so expensive runs delay the benefit
of its learned surrogate~\citep{falkner2018bohb}. Parallel variants can choose new
configurations while evaluations are running, but must account for their unknown outcomes,
for example through Constant Liar~\citep{ginsbourger2010kriging}. Freeze-thaw
Bayesian optimization~\citep{swersky2014freeze} combines learning-curve prediction with
iterative resource allocation for modern deep-learning workloads. Predicting how partial curves
continue provides earlier feedback, allows concurrent trials, and lets the optimizer revisit runs
whose early performance was weak.

DyHPO~\citep{wistuba2022dyhpo} models configurations jointly with their partial learning
curves. Prior-data fitted networks (PFNs) are transformers trained on synthetic tasks drawn
from a prior. At inference time, they use a single forward pass to approximate predictions
under that prior~\citep{muller2022pfn}. LC-PFN first
applied them to extrapolating individual learning curves~\mbox{\citep{adriaensen2023lcpfn}}.
FT-PFN~\citep{rakotoarison2024ifbo} extended this approach to freeze--thaw optimization by
conditioning jointly on hyperparameter configurations and their partial curves, without
per-task surrogate fitting.

FT-PFN pads every hyperparameter vector to a fixed length and projects it with a linear layer.
Each input position has its own weights, so increasing the maximum number of hyperparameters
changes the layer and requires retraining, while shorter configurations leave positions unused.
ICICLE instead forms one token per hyperparameter using shared operations and treats the tokens
as a set. Feature self-attention lets the hyperparameters interact, and attention pooling reduces
the variable-size set to one configuration embedding. Because no parameter shape depends on the
number or ordering of hyperparameters, the same encoder handles spaces of different sizes.

In this work, we:
\begin{itemize}
  \item introduce a factored, width-free surrogate for freeze--thaw optimization that supports
        variable-size hyperparameter spaces;
  \item evaluate this architecture together with the ICICLE prior, an extension of FT-PFN's
        BNN-based learning-curve prior that covers typed hyperparameters and a broader range
        of noise and learning-curve behavior; and
  \item provide a JAX implementation that samples synthetic tasks on the accelerator rather
        than precomputing a fixed training set, together with a freeze--thaw scheduler for
        Ray Tune and ONNX export for inference without JAX or PyTorch.
\end{itemize}
We evaluate ICICLE under matched protocols across LCBench, PD1, TaskSet,
Meta-Album, and ARLBench.\footnote{Code and pretrained model: \url{https://github.com/lauritsf/icicle}.}

\begin{figure}[t]
  \centering
  \includegraphics[width=\textwidth]{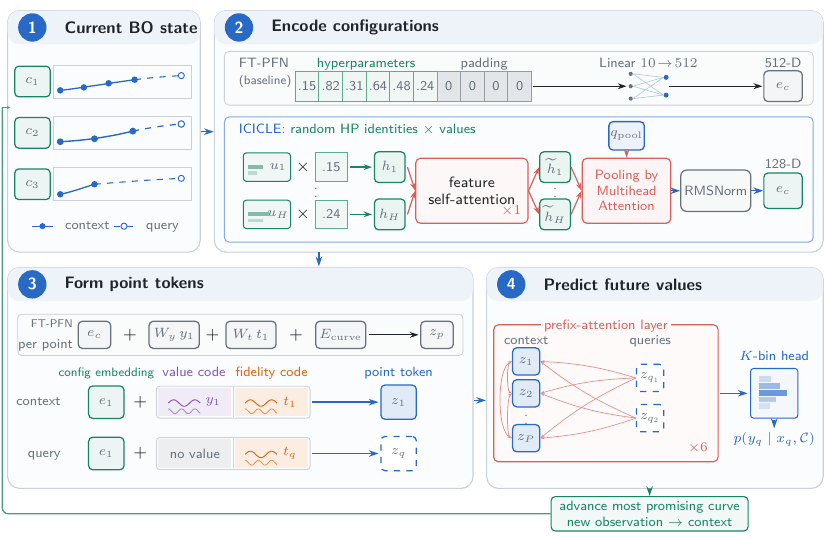}
  \caption{ICICLE's freeze--thaw Bayesian-optimization loop. (1) Observed curve prefixes
  form the context; later fidelities are queries. (2) Feature self-attention and attention
  pooling encode each configuration once, using one value--identity token per hyperparameter.
  (3) Observations combine the gathered configuration embedding with separate value and
  fidelity encodings; queries omit the value encoding. (4) Each query attends independently
  to the context, and a 1{,}000-bin head predicts its value distribution. The optimizer
  advances a configuration and adds the observation to the context.
  Appendix~\ref{app:surrogate-details} gives the architecture details.}
  \label{fig:architecture}
\end{figure}

\section{Method}\label{sec:method}

\subsection{Factored configuration--fidelity surrogate}\label{sec:arch}
Inspired by the Set Transformer~\citep{lee2019settransformer}, ICICLE represents a configuration
$c=(v_1,\ldots,v_H)$ as the set of value--identity tokens $\{v_j u_j\}_{j=1}^{H}$, where $v_j$
is a centered, variance-normalized hyperparameter value and $u_j$ is its task-random identity
vector. The identity vectors are orthogonal within a task and redrawn for every task, so the
model can distinguish the hyperparameters without assigning a fixed meaning to a position.
Feature self-attention represents interactions among the tokens, and Pooling by Multihead
Attention (PMA) maps the variable-size set to one configuration embedding $e_c$. Because no
learned weight shape depends on $H$, the same encoder handles different numbers of
hyperparameters without padding. ICICLE computes $e_c$ once and reuses it at every point on the
configuration's curve. Each point combines this embedding with separate fixed sinusoidal
encodings of observed performance and fidelity; query tokens omit observed performance. FT-PFN
instead applies a linear layer to a rescaled, zero-padded ten-position vector at every point.
Appendix~\ref{app:surrogate-details} gives the complete architecture specification.

ICICLE and FT-PFN both perform in-context curve extrapolation without task-specific fitting,
using observed curve prefixes as context and later fidelities as queries. Both models use six
prefix-attention layers, the same masking pattern, and a 1{,}000-bin output head to produce
predictive distributions in a single forward pass. ICICLE uses an embedding width of 128
rather than FT-PFN's 512, and a feed-forward width of 256 rather than 1{,}024. This yields
1.28M parameters, 8.7\% of FT-PFN's 14.7M. This smaller size appears sufficient on the
benchmarks studied.

\subsection{The ICICLE training prior}\label{sec:prior}
The ICICLE prior builds on FT-PFN's BNN-based learning-curve prior and retains its BNN
core~\citep{rakotoarison2024ifbo}. We sample four to eight BNN layers rather than FT-PFN's
eight to fifteen so that differences between configurations remain more strongly reflected
in the generated curves. The extensions introduce continuous, ordinal, and nominal
hyperparameters, Student-t observation noise, quantized values, and configuration-dependent
divergence.

It samples each hyperparameter as continuous, ordinal, or nominal, with types
and level counts redrawn per task. Ordinal levels are evenly spaced on $[0,1]$, while
nominal levels receive independent random values, so category indices do not impose an
ordering.

The prior uses Student-t observation noise with degrees of freedom and scale sampled
per task, and quantizes values to a grid with task-specific resolution. Divergence depends
on configuration: affected curves collapse during a sampled window that may span the run,
begin after training starts, or end in recovery.

Appendix Table~\ref{tab:prior-spec} gives the complete settings and corresponding FT-PFN values.

\section{Experimental setup}\label{sec:setup}

On LCBench, PD1, and TaskSet, we compare ICICLE with the
published FT-PFN model over a 1{,}000-iteration horizon. Both methods run twenty BO
seeds (0--19) on paired benchmark
scenarios~\citep{picard2021seed}. Each benchmark is evaluated separately, and tasks receive
equal weight. ICICLE and FT-PFN use the same MFPI-random acquisition.

We separately compare ICICLE with Quick-Tune~\citep{arango2024quicktune}, ASHA, and random
search on 86 Meta-Album
tasks with 33 hyperparameters. We use the same ICICLE model, without any pre-training or
meta-training on Meta-Album tasks. Quick-Tune uses its meta-learned performance and cost
predictors. All four
methods share candidate pools and paired initial configurations, and each runs twenty BO seeds
(0--19). ICICLE adopts Quick-Tune's
cost-aware acquisition principle while retaining MFPI's exceedance score, and estimates its
value map from observations rather than requiring a prespecified performance range. The budget
axis counts table training plus evaluation time and is capped at 1, 4, and 16 hours for micro,
mini, and extended. Appendix~\ref{app:protocol-qtmeta} gives the complete protocol; method
overhead is reported in Appendix~\ref{app:protocol-overhead}.

\section{Results}\label{sec:results}

\begin{figure}[t]
  \centering
  \includegraphics[width=\textwidth]{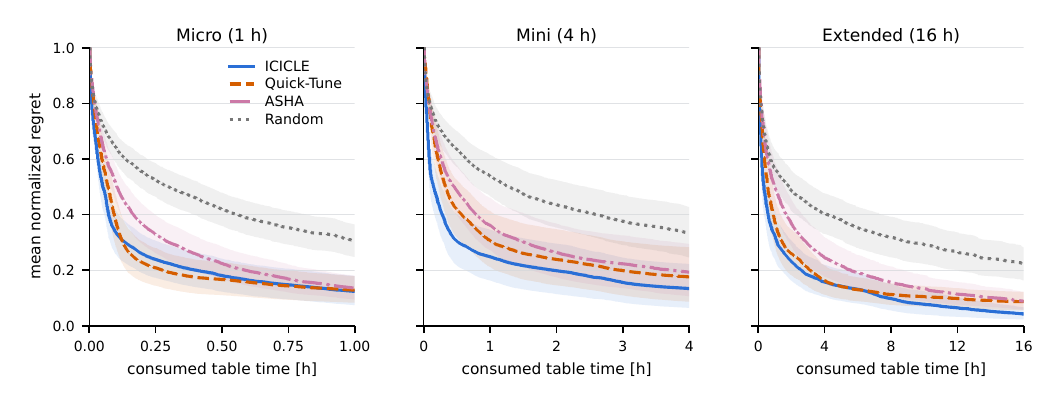}
  \caption{Mean normalized regret versus consumed benchmark-table time on Meta-Album
(86 tasks). Curves average 20 BO seeds per task; ribbons show pointwise
95\% bootstrap confidence intervals.}
  \label{fig:qtmeta-v2}
\end{figure}

ICICLE reaches lower mean final normalized regret than Quick-Tune overall
(0.103 versus 0.133; 95\% CI for the paired reduction: [0.006, 0.056]).
The difference is clearest on extended; mini and micro intervals include zero.
Appendix~\ref{app:qtmeta-tasks} shows the task-level results.

\begin{figure}[t]
  \centering
  \includegraphics[width=\textwidth]{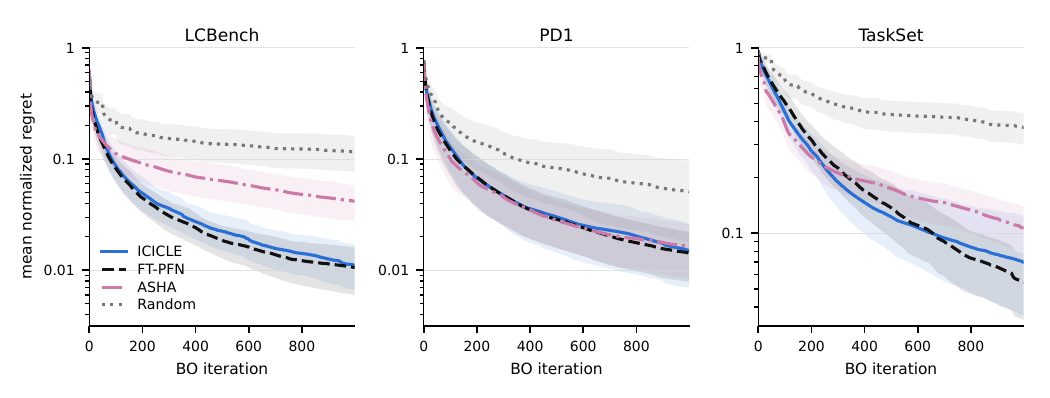}
  \caption{Mean normalized regret on LCBench (35 tasks), PD1 (28), and TaskSet (24).
Curves average 20 BO seeds per task; ribbons show pointwise 95\% bootstrap
confidence intervals.}
  \label{fig:ftpfn-regret}
\end{figure}

ICICLE and FT-PFN follow similar mean-regret trajectories across LCBench, PD1, and TaskSet.
Both remain below random search; ASHA is closest on PD1. Appendix~\ref{app:ftpfn-views}
examines the adaptive value map and compares training seeds 0--2 for ICICLE and
width-128 FT-PFN.

\section{Conclusion and limitations}\label{sec:discussion}

ICICLE improves on Quick-Tune overall and on extended without Meta-Album training;
mini and micro show no clear difference. It produces similar trajectories to FT-PFN with 8.7\% of its
parameters, though low endpoint regret limits resolution. ICICLE delivers competitive freeze--thaw optimization across these benchmarks
with a compact surrogate trained entirely on synthetic tasks.

\clearpage
\section*{Acknowledgments}

The authors acknowledge support from the Novo Nordisk Foundation under grant no NNF22OC0076658
(Bayesian neural networks for molecular discovery).

We acknowledge the Danish e-infrastructure Consortium (DeiC) for awarding this project access to the
LUMI supercomputer, owned by the EuroHPC Joint Undertaking, hosted by CSC (Finland) and the LUMI
consortium.

\bibliographystyle{plainnat}
\bibliography{references}

\appendix

\section{Protocol details}\label{app:protocol}

Throughout the main comparisons, ICICLE uses training seed 0, FT-PFN uses its published
checkpoint, and Quick-Tune uses meta-training seed 0 within each held-out fold.

\subsection{Meta-Album}\label{app:protocol-qtmeta}
We evaluate ICICLE, Quick-Tune, ASHA, and random search on the 86 Meta-Album tasks for which
the Quick-Tune release~\citep{arango2024quicktune} provides tables: 30 micro, 30 mini, and
26 extended. ICICLE receives no pre-training or meta-training on Meta-Album tasks and uses
the cost-aware acquisition described in Appendix~\ref{app:optimization-details}. Quick-Tune
uses its native expected-improvement-over-predicted-cost acquisition. For each split, we use
Quick-Tune's official five-fold division. Its performance and cost predictors are meta-trained
on the tasks outside the held-out fold and evaluated only on tasks within that fold. We train
five meta-training seeds for every fold and split, giving $3\times5\times5=75$ trainings.
ASHA uses reduction factor $\eta=3$ and grace period one. Every method is evaluated with BO
seeds 0--19.

For each task and BO seed, all methods use the same candidate table, lexicographically indexed
candidate pool, and initial configuration. Subsequent proposals follow each method's own policy.
An accepted action advances one configuration by its next valid epoch and charges the
corresponding table-reported training and evaluation time. The first proposal that would exceed
the budget is rejected without revealing its objective. The budgets are 1, 4, and 16 hours for
micro, mini, and extended, respectively.

Let $f_{\min}$ and $f_{\max}$ be the smallest and largest finite objective values in a task
table, and let $f_t$ be the best value observed by time $t$. We report normalized regret
\[
r_t=\frac{f_{\max}-f_t}{f_{\max}-f_{\min}},
\]
clipped to $[0,1]$. For each displayed trained model, BO replicas are averaged within each
task before tasks receive equal weight. Curves show the task mean with pointwise 95\% bootstrap confidence intervals,
using the two-level procedure described below separately within each split.

\subsection{Meta-Album computational overhead}\label{app:protocol-overhead}
The horizontal axis in Figure~\ref{fig:qtmeta-v2} counts benchmark-table training and
evaluation time, excluding optimization time. We measure overhead across all 86 tasks
and 20 BO seeds per task. After compilation, ICICLE batches 128 trajectories on micro
and 32 on mini and extended. We time synchronized device execution, including fused
table lookup, and divide total time by the number of trajectories to report amortized
cost per run. Host input preparation and result conversion are excluded. Quick-Tune
runs sequentially; its recorded suggestion and observation/refitting times exclude
table lookup and serialization.
ICICLE's amortized overhead is 0.74, 2.59, and 2.95 seconds per run on micro,
mini, and extended, respectively (at most 0.021\% of each split's budget).
Quick-Tune's median overhead is 6.7\%, 1.3\%, and 0.3\% of the respective budgets.

\subsection{LCBench, PD1, and TaskSet}\label{app:protocol-ftpfn}
We compare ICICLE with the published FT-PFN checkpoint on LCBench (35 tasks), PD1
(28 task-model instances), and TaskSet (24 task-optimizer instances). Both surrogate methods
use MFPI-random (Appendix~\ref{app:optimization-details}), while ASHA and random search retain
their own allocation rules. Each trajectory contains an initial observation followed by 999
allocation steps. Every method uses BO seeds 0--19, which are averaged within each task.

We use the 28 PD1 benchmark instances listed in the ifBO experiment repository,
corresponding to the 16 task--model--batch-size combinations described in their
paper~\citep{rakotoarison2024ifbo}, with multiple fidelity subsampling levels for four
combinations.

Let $g_c=\max_\ell y_c(\ell)$ be the best value attained by configuration $c$ over all
fidelities, and let $f_t$ be the best value observed by iteration $t$. We report
\[
r_t=\frac{\max_c g_c-f_t}{\max_c g_c-\min_c g_c},
\]
clipped to $[0,1]$. The benchmark adapters express objectives as higher-is-better values in
$[0,1]$. The published FT-PFN checkpoint receives these values directly. ICICLE applies its
adaptive map to the observations and incumbent and evaluates acquisition in the mapped
model units (Appendix~\ref{app:optimization-details}). Regret is computed from the stored
adapter-scale observations, independently of this map.

Within each task and BO seed, ICICLE and FT-PFN receive the same initial observation and
MFPI-random draws. BO replicas are averaged within each task before tasks receive equal
weight; trained checkpoints are kept separate. Each benchmark is reported separately.

\paragraph{Uncertainty.}
We average the 20 BO runs within each task and then give each task equal weight.
Ribbons show pointwise 95\% bootstrap confidence intervals, computed by resampling
tasks and then BO runs within each task 20{,}000 times. Each run contributes its full
trajectory, and paired comparisons use the same samples for both methods. Each
trained model is evaluated separately. For Meta-Album, we compute intervals within
each split and use all 86 tasks for the overall comparison.

\subsection{ARLBench}\label{app:protocol-arlbench}
We additionally report a table-based evaluation on 42 ARLBench tasks
\citep{becktepe2026arlbench}. For each task, configuration, and fidelity, the objective is
averaged over the ten provided training seeds. The search spaces contain 10--13
hyperparameters. We compare ICICLE with ASHA, Hyperband, and random search, using the same
1{,}000-iteration horizon, BO seeds, fixed-checkpoint selection, and normalized-regret definition
as above. A crashed or non-finite table entry is assigned the smallest finite performance in
that task. These results use benchmark tables; evaluation on the live benchmark remains future
work.

\section{Method details}\label{app:method}

\subsection{Training prior}\label{app:prior-details}

\begin{table}[H]
  \caption{Differences between FT-PFN's synthetic training prior and the ICICLE prior.
  $\mathrm{LogU}(a,b)$ denotes a log-uniform draw, and
  $\mathrm{TP}(\lambda;0{:}m)$ a Poisson draw conditioned on the displayed support.}
  \label{tab:prior-spec}
  \centering
  \footnotesize
  \setlength{\tabcolsep}{4pt}
  \begin{tabular}{@{}>{\raggedright\arraybackslash}p{0.16\textwidth}
                      >{\raggedright\arraybackslash}p{0.29\textwidth}
                      >{\raggedright\arraybackslash}p{0.47\textwidth}@{}}
    \toprule
    Component & FT-PFN & ICICLE prior \\
    \midrule
    BNN depth
      & $L \sim U\{8,\ldots,15\}$
      & $L \sim U\{4,\ldots,8\}$ \\
    BNN core
      & Published weight scale, pre-activation and output noise, and 0.145 DropConnect
      & Unchanged \\
    Active HP count
      & $d \sim U\{1,\ldots,10\}$
      & $\lambda_d \sim \mathrm{LogU}(1,20)$;
        $d=1+\mathrm{TP}(\lambda_d;0{:}31)$ \\
    HP types
      & All active slots continuous
      & Per-task rates $\lambda_c \sim \mathrm{LogU}(1,20)$,
        $\lambda_o \sim \mathrm{LogU}(0.05,10)$, and
        $\lambda_n \sim \mathrm{LogU}(0.05,4)$; each slot draws continuous, ordinal, or
        nominal in proportion to these rates \\
    Discrete levels
      & Not applicable
      & $\lambda_k \sim \mathrm{LogU}(0.3,8)$;
        $k=2+\mathrm{TP}(\lambda_k;0{:}22)$. Ordinal values are evenly spaced on $[0,1]$;
        nominal values are i.i.d. $U(0,1)$ \\
    Observation noise
      & Clipped Gaussian with per-configuration
        $\sigma_c \sim \mathrm{LogNormal}(-5,1)$
      & Clipped, variance-standardized Student-t with
        $\nu \sim \mathrm{LogU}(2.1,8)$; the same $\sigma_c$ is multiplied by a per-task
        $s \sim \mathrm{LogU}(0.01,8)$ \\
    Quantization
      & None
      & $N=\mathrm{round}(\mathrm{LogU}(10^2,3\!\times\!10^4))$;
        $y \leftarrow \mathrm{round}(Ny)/N$ after clipping \\
    Divergence
      & None
      & A per-task fraction $q \sim \mathrm{LogU}(0.02,0.6)$; BNN outputs select the
        affected configurations and their alive windows, outside which values are clamped
        and noise is removed \\
    Query schedule
      & Only unobserved steps are eligible as queries
      & Each already-observed step is also eligible with probability 0.75; its target is the
        same stored observation \\
    \bottomrule
  \end{tabular}
\end{table}

For an affected curve, the two divergence-window constructions are equally likely. A crash
window begins alive and draws $t_{\mathrm{die}} \sim U(-0.25,1)$. A recovery window draws
$t_{\mathrm{alive}} \sim U(0,1.2)$ and, with equal probability, either remains alive or
crashes again before the end. Outside the alive window, the clamp is zero with probability
0.5 and otherwise uniform between zero and the curve's initial value.

\subsection{Surrogate architecture}\label{app:surrogate-details}

\paragraph{Configuration encoder.}
ICICLE encodes each configuration once. Every centered, variance-normalized hyperparameter
value scales a task-random identity tag, one feature self-attention layer mixes the resulting
tokens, and PMA~\citep{lee2019settransformer} with RMSNorm reduces them to a single embedding
$e_c$. During training, each task receives a fresh random orthonormal frame: we draw an
$E\times H$ matrix with independent standard-normal
entries, where $E$ is the embedding width and $H$ the number of hyperparameters, apply
sign-corrected QR orthogonalization, and scale each identity vector to norm $\sqrt{E/2}$.
Evaluation uses a fixed draw. The encoding is independent of fidelity and observed value, so
the same $e_c$ is gathered into
every point of the curve; no weight shape depends on the number of hyperparameters. The current
orthonormal-tag construction requires the hyperparameter count not to exceed the embedding
width, so the released model supports at most 128 hyperparameters. FT-PFN instead rescales the
$k$ active values by $10/k$, zero-pads to ten slots, and applies a linear $10\!\to\!512$
projection at every point.

\paragraph{Point tokens.}
To bind fidelity and observed value to the configuration encoding, we add sinusoidal codes to the
gathered $e_c$: Fourier features of the value in $[0,1]$ with frequencies $10000^{-2i/d}$,
so the top channel spans one cycle across the range and the base only sets how far down the
ladder reaches. Value and fidelity occupy disjoint 64-channel halves. A query token gets no
value code. Each point uses the shared encoding of its configuration. In FT-PFN the
value and budget enter as learned linear projections and the curve as a learned id
embedding with room for 1{,}000 curves; a query's value contribution is zeroed.

\paragraph{Transformer and head.}
The point tokens enter a six-layer prefix-attention transformer: context tokens attend
within the context, and each query attends to the context but not to other queries, so
predictions are independent given the observations. A bar-distribution head over 1{,}000
uniform bins on $[0,1]$ maps each query's output to a predictive distribution. FT-PFN
shares this depth, masking, and head; the difference is width, 128 against 512 (FFN 256
against 1{,}024), which is where the parameter gap comes from: 1.28M against 14.7M.

\subsection{Training recipe}\label{app:training-details}
We implement training in JAX, including a JIT-compiled prior sampler that generates fresh
synthetic tasks at every update. This avoids precomputing and storing a training corpus;
FT-PFN's released training recipe instead loads a pre-generated prior dataset from disk.
We train three independent ICICLE models with seeds 0--2. Each model is trained for 80{,}000 updates
with batches of 128 tasks and sequences of 1{,}000 points, minimizing the mean bar-distribution
NLL over query points. This
amounts to 10.24 million synthetic task draws per model, compared with approximately 2 million
for the published FT-PFN checkpoint. We use AdamW with zero weight decay, $\beta_1=0.9$,
$\beta_2=0.999$, $\epsilon=10^{-8}$, and global gradient clipping at 1.0. The learning rate
warms linearly to $2.4\times10^{-3}$ over 4{,}000 updates, remains constant for 68{,}000, and
decays linearly to zero over the final 8{,}000. Computation uses bfloat16 while parameters and
the loss remain in float32, with configuration-encoder activation recomputation. We record validation NLL
every 2{,}000 updates on a fixed set of 128 synthetic tasks. The reported models are the resulting
step-79{,}999 checkpoints.

\subsection{Optimization}\label{app:optimization-details}

\paragraph{Acquisition.}
On LCBench, PD1, TaskSet, and ARLBench, ICICLE uses MFPI-random, the acquisition used with
FT-PFN~\citep{rakotoarison2024ifbo}.
Let $n_i$ be the number of observed levels for configuration $i$, $b_i$ the length of its
curve, and
\[
q_i(h)=\min\{n_i+h,b_i\}.
\]
At iteration $t$, we draw
\[
h\sim U\{1,\ldots,b_{\max}\},\qquad
\log_{10}\tau\sim U(-4,-1).
\]
Let $g_t$ be the value map used by the surrogate (the identity for published
FT-PFN). Write $z_i=g_t(y_i)$ and $z_{\mathrm{best}}=g_t(f_{\mathrm{best}})$,
and set
\[
T_t=z_{\mathrm{best}}+\tau(1-z_{\mathrm{best}}).
\]
Each configuration is scored by
\[
a_i(h,T_t)
=\Pr\!\left(z_i(q_i(h))>T_t\mid\mathcal D_t\right),
\]
where $\mathcal D_t$ contains the observations available at iteration $t$. The configuration
with the highest score is advanced by one level.

On Meta-Album, we retain the MFPI score but adopt the cost-aware principle used by
Quick-Tune~\citep{arango2024quicktune}. The score becomes
\[
\widetilde a_i(h,T_t)
=
\frac{a_i(h,T_t)}
{\left(\Delta c_i(h)+c_{\min}\right)^{\alpha_t}},
\qquad
\alpha_t=
\mathrm{clip}\!\left(\frac{B-C_t}{B},0,1\right).
\]
Here, $c_i(k)$ is cumulative table training and evaluation time through level $k$,
divided by the split budget and clipped to $[0,1]$.
Then $\Delta c_i(h)=c_i(q_i(h))-c_i(n_i)$ is the cost of reaching the queried fidelity,
$c_{\min}$ is the smallest positive increment of $c_i$, $C_t$ is the cost consumed so far,
and $B$ is the total cost budget. The exponent
decreases from one to zero as the budget is consumed, making the acquisition cost-sensitive
early in the run and progressively less cost-sensitive near the budget limit. Quick-Tune
divides expected improvement by predicted cost; our implementation applies the same
cost-aware principle to MFPI's exceedance probability.

\paragraph{Adaptive value map.}
The benchmark adapters express curve values on a higher-is-better $[0,1]$ scale, which the
published FT-PFN checkpoint receives directly. Before each decision, ICICLE affinely maps
the observed 5th percentile and maximum to $1/3$ and $2/3$, leaving room for predictions
above and below this interval. Mapped values are clipped to $[0,1]$. When the anchor range
is numerically degenerate, observations and the incumbent map to $1/2$.
The acquisition threshold and exceedance probability are computed in mapped
units. Reported regret is computed from the stored benchmark-adapter values using
the benchmark-specific normalization in Appendices~\ref{app:protocol-qtmeta}
and~\ref{app:protocol-ftpfn}.

\clearpage
\section{Task-level Quick-Tune results}\label{app:qtmeta-tasks}

\begin{figure}[!ht]
Across all three splits, tasks are grouped by held-out Meta-Album set and ordered within
each set by their mean normalized regret across methods and BO seeds. Large markers show
per-task means across 20 BO seeds; small markers show individual BO runs.
ASHA and random search contribute one run per BO seed.

  \centering
  \includegraphics[width=\textwidth,height=0.82\textheight,keepaspectratio]{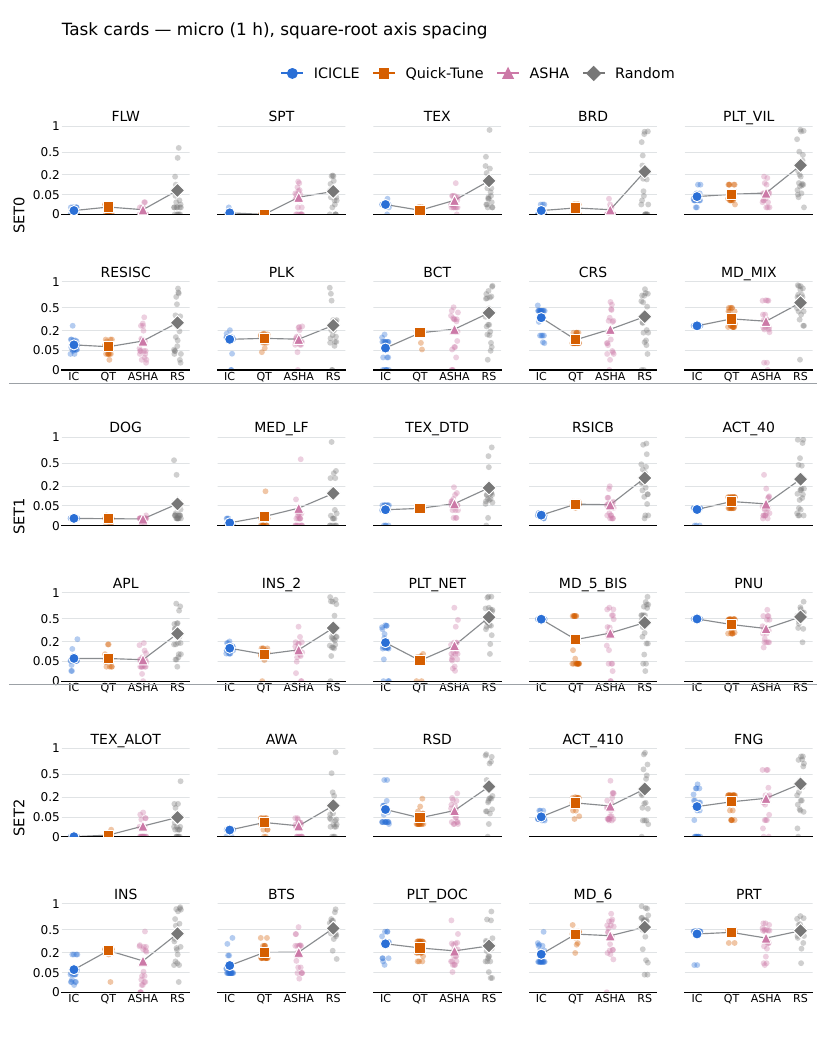}
  \caption{Task-level final normalized regret on the micro split (1 h; 30 tasks). Lower is
  better.}
  \label{fig:qtmeta-tasks-micro}
\end{figure}

\clearpage
\begin{figure}[t]
  \centering
  \includegraphics[width=\textwidth,height=0.82\textheight,keepaspectratio]{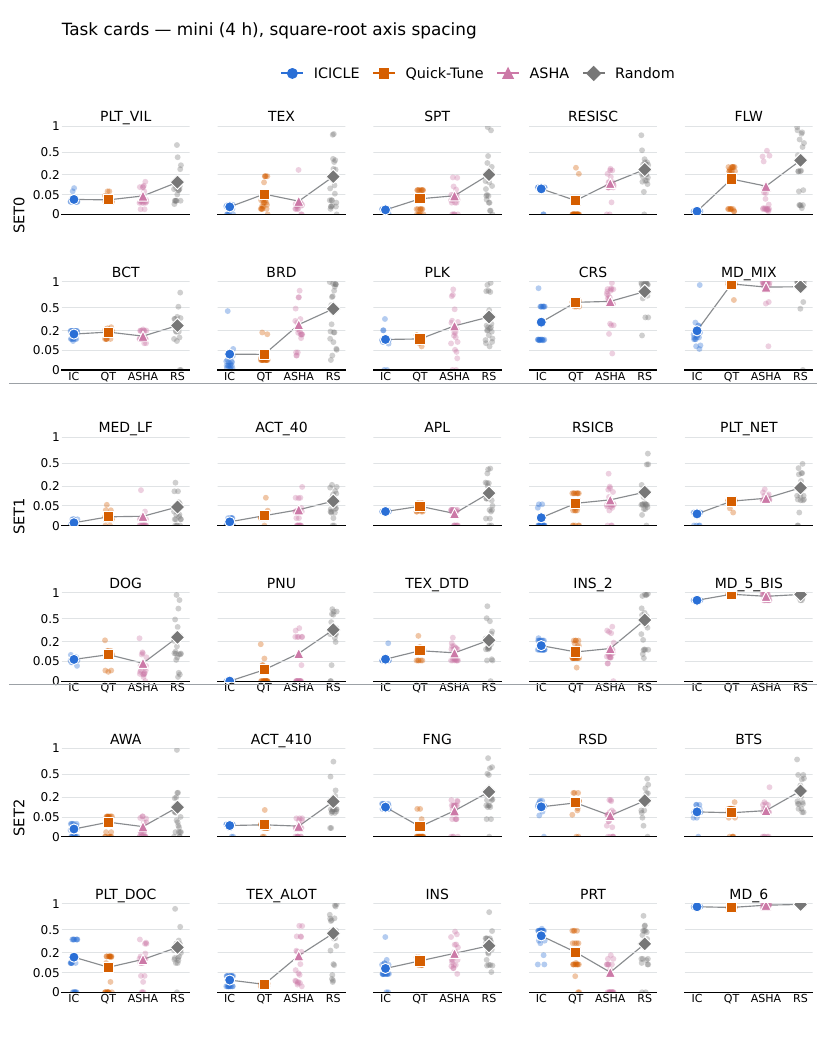}
  \caption{Task-level final normalized regret on the mini split (4 h; 30 tasks). Lower is
  better.}
  \label{fig:qtmeta-tasks-mini}
\end{figure}

\clearpage
\begin{figure}[t]
  \centering
  \includegraphics[width=\textwidth,height=0.82\textheight,keepaspectratio]{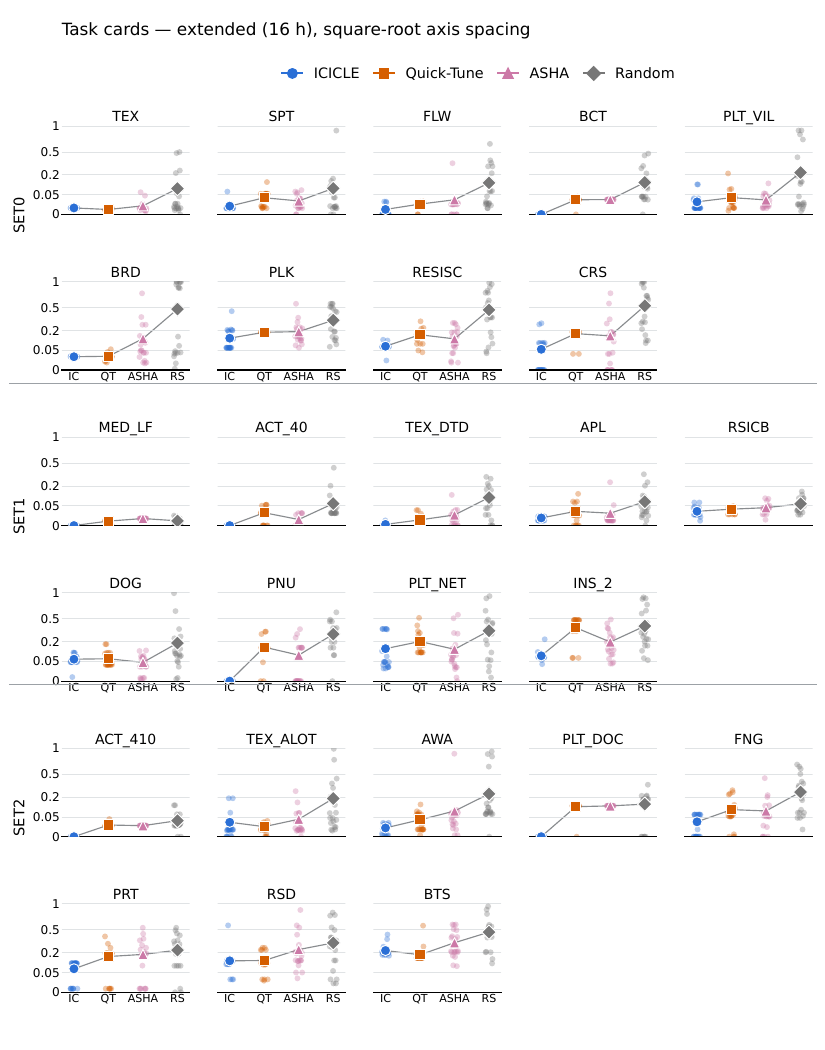}
  \caption{Task-level final normalized regret on the extended split (16 h; 26 tasks). Lower
  is better.}
  \label{fig:qtmeta-tasks-extended}
\end{figure}

\clearpage
\section{Additional FT-PFN benchmark views}\label{app:ftpfn-views}

While the published FT-PFN model has width 512 and accepts up to 10
hyperparameters, we trained width-128 models accepting up to 33 hyperparameters
on our synthetic prior, using training seeds 0--2. We trained each model for
80{,}000 updates and evaluated it with ICICLE's adaptive value map.
We found that FT-PFN also works well with a smaller network and more
hyperparameter inputs. Our width-128 models perform similarly to ICICLE on
Meta-Album, but tend to fall behind both ICICLE and the published FT-PFN model
on LCBench and TaskSet.

\begin{figure}[!ht]
  \centering
  \includegraphics[width=\textwidth]{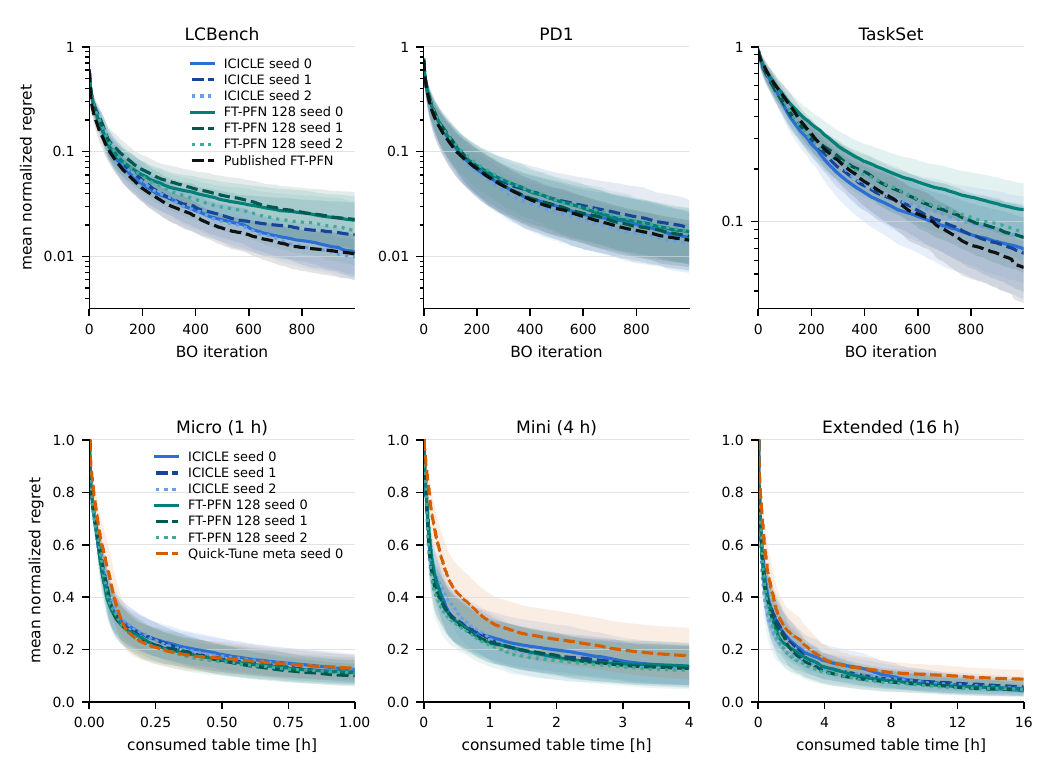}
  \caption{Training-seed variation for ICICLE and our trained FT-PFN-128 variants (seeds 0--2).
The upper row shows the core benchmarks with published FT-PFN as a reference;
the lower row shows Meta-Album with Quick-Tune meta seed 0. Each curve uses
20 BO seeds per task; ribbons show pointwise 95\% bootstrap confidence intervals.}
  \label{fig:training-seeds}
\end{figure}

\begin{figure}[!ht]
  \centering
  \includegraphics[width=\textwidth]{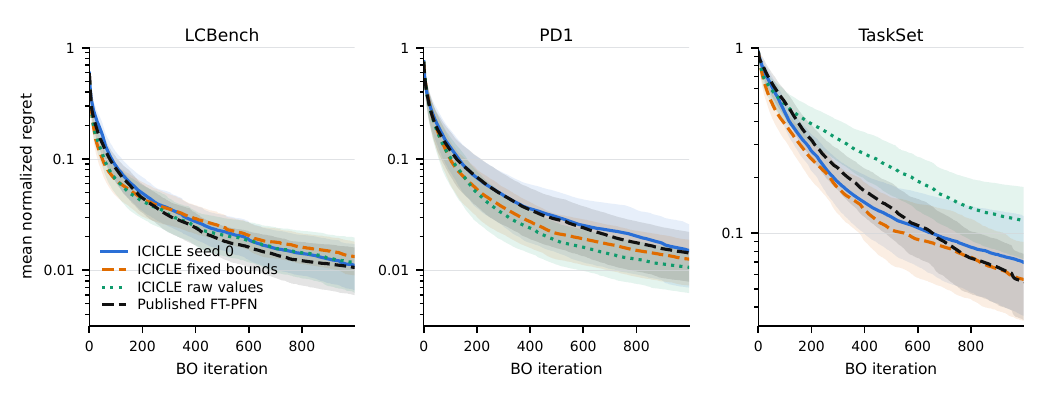}
  \caption{Value-presentation ablation on LCBench, PD1, and TaskSet. ICICLE uses the adaptive
  map described in Appendix~\ref{app:optimization-details}. ``Raw values'' passes the
  benchmark-adapter values in $[0,1]$ to ICICLE unchanged, while ``fixed bounds'' maps the
  complete $[0,1]$ interval to $[\nicefrac{1}{3},\nicefrac{2}{3}]$. FT-PFN receives the
  benchmark-adapter values directly.}
  \label{fig:ftpfn-presentations}
\end{figure}

\clearpage
\begin{figure}[t]
In the task-level views, large markers show method means across 20 BO seeds;
small markers show individual BO runs. Baseline methods contribute one run per BO seed.

  \centering
  \includegraphics[width=\textwidth,height=0.82\textheight,keepaspectratio]{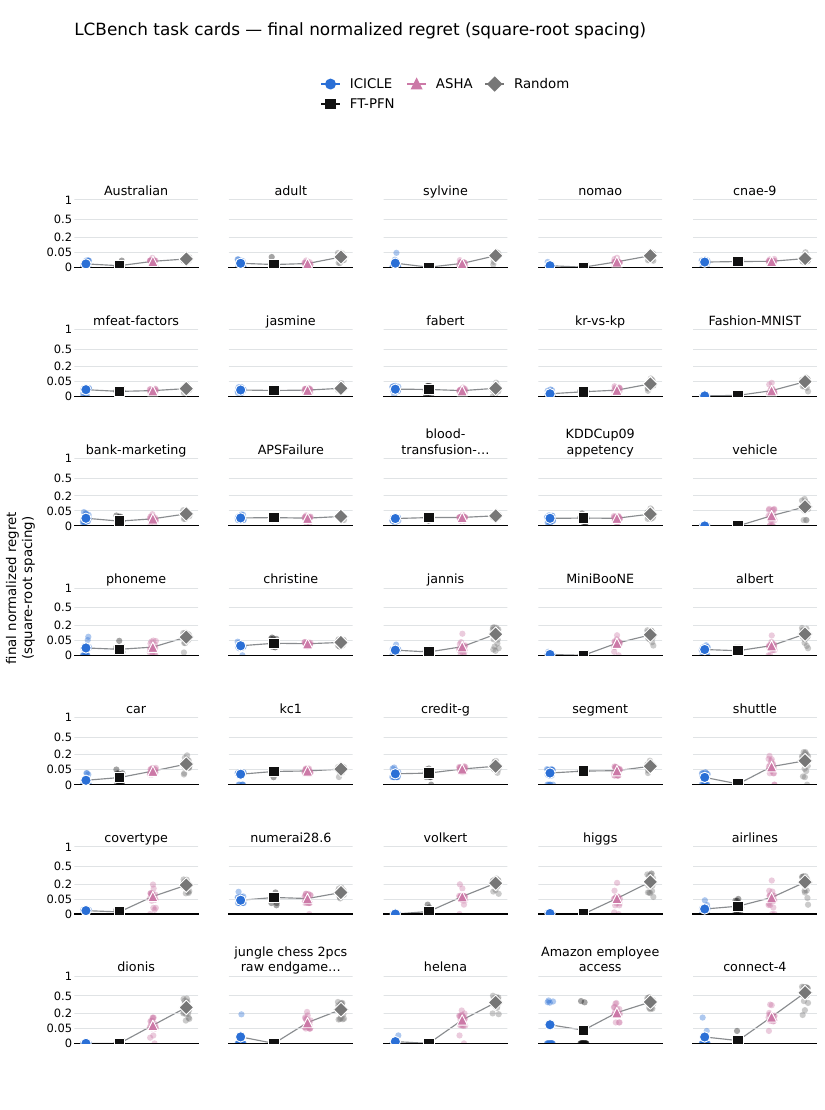}
  \caption{Task-level normalized regret on 35 LCBench tasks after a 1{,}000-iteration
  horizon.}
  \label{fig:ftpfn-tasks-lcbench}
\end{figure}

\clearpage
\begin{figure}[t]
  \centering
  \includegraphics[width=\textwidth,height=0.82\textheight,keepaspectratio]{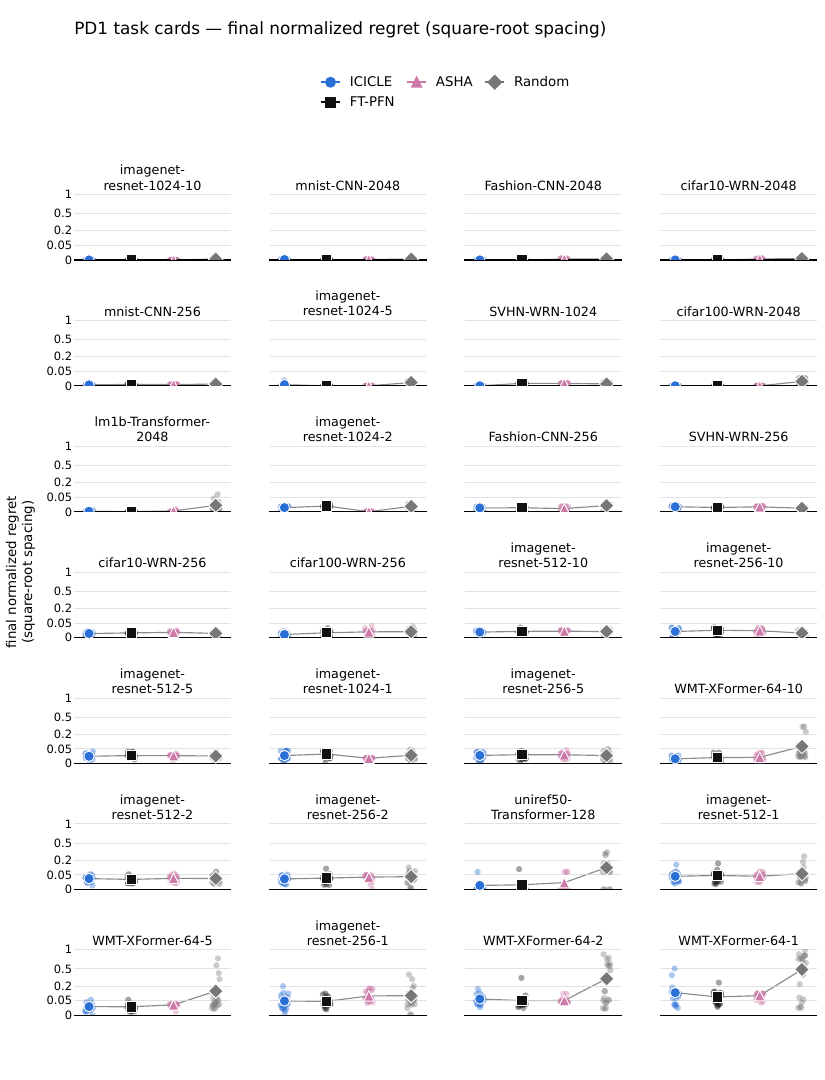}
  \caption{Task-level normalized regret on 28 PD1 tasks after a 1{,}000-iteration
  horizon.}
  \label{fig:ftpfn-tasks-pd1}
\end{figure}

\clearpage
\begin{figure}[t]
  \centering
  \includegraphics[width=\textwidth,height=0.82\textheight,keepaspectratio]{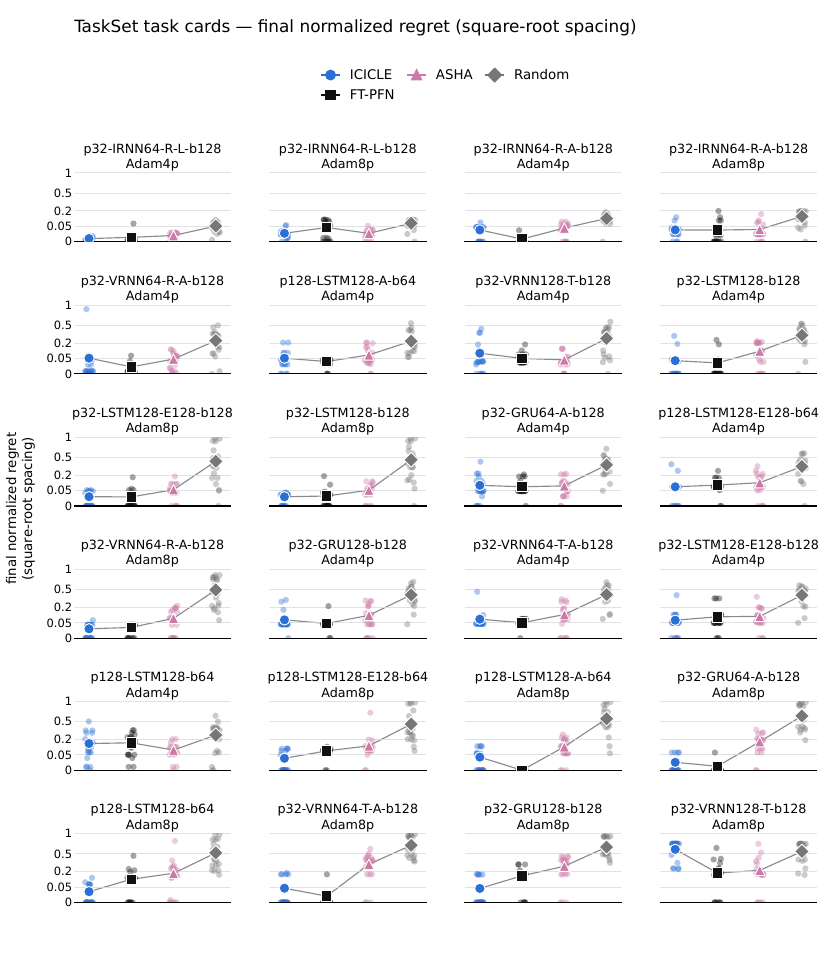}
  \caption{Task-level normalized regret on 24 TaskSet tasks after a 1{,}000-iteration
  horizon.}
  \label{fig:ftpfn-tasks-taskset}
\end{figure}

\clearpage
\section{Table-based ARLBench results}\label{app:arlbench-tasks}

\begin{figure}[!ht]
  \centering
  \includegraphics[width=\textwidth]{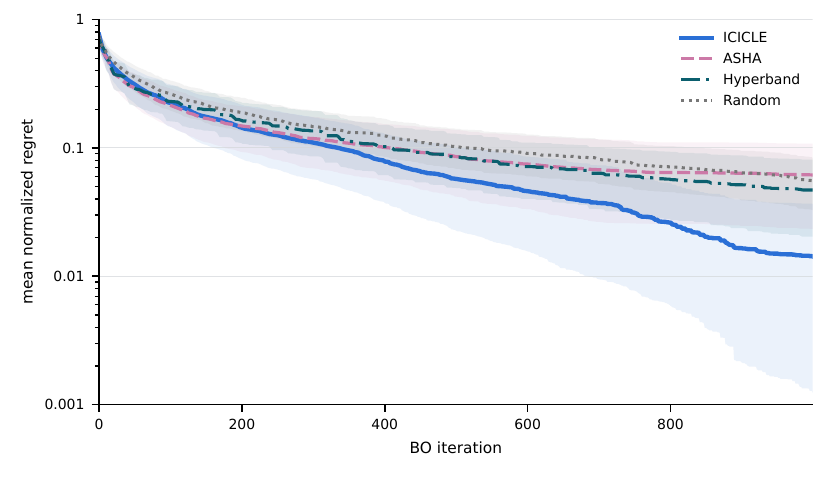}
  \caption{Table-based freeze--thaw optimization on 42 ARLBench mean-seed tasks. Mean
  normalized regret (lower is better) is shown over 1{,}000 iterations; ribbons are pointwise
  95\% bootstrap confidence intervals over tasks and BO runs. Each method uses
  20 BO seeds per task. Crash outcomes follow ARLBench's minimum-performance
  convention. These results are not directly comparable to evaluations on the live benchmark.}
  \label{fig:arlbench-mean}
\end{figure}

\clearpage
\begin{figure}[t]
In the task-level view, tasks are ordered by their mean normalized regret across methods and
BO seeds. Large markers show per-task means across 20 BO seeds; small markers show individual
BO runs. Baseline methods contribute one run per BO seed.

  \centering
  \includegraphics[width=\textwidth,height=0.82\textheight,keepaspectratio]{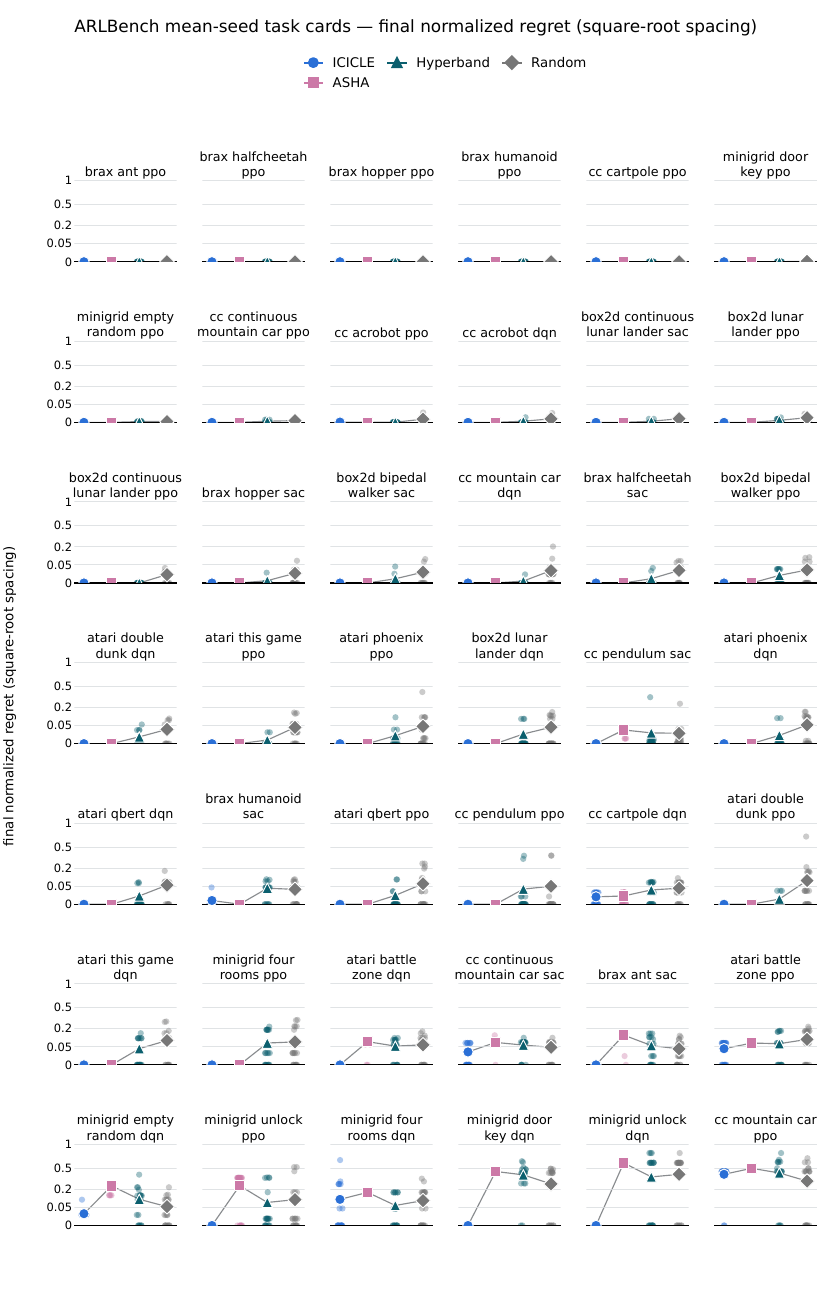}
  \caption{Task-level final normalized regret on the 42 ARLBench mean-seed tasks after a
  1{,}000-iteration horizon. Lower is better.}
  \label{fig:arlbench-tasks}
\end{figure}

\end{document}